\documentclass{article}

\usepackage{microtype}
\usepackage{graphicx}
\usepackage{amsmath,amssymb,amsfonts}
\usepackage{array}
\usepackage{booktabs} 
\usepackage{amsfonts} 
\usepackage[subrefformat=parens,farskip=0pt,justification=centering]{subfig}
\usepackage[ruled,vlined]{algorithm2e}
\usepackage[table]{xcolor}
\usepackage{float}
\usepackage{multirow}
\usepackage{mathtools}
\usepackage[title]{appendix}

\usepackage{hyperref}

\newtheorem{problem}{Problem}

\usepackage[accepted]{sysml2019}

\def\gL{{\mathcal{L}}}
\def\HD{{\mathrm{HD}}}
\def\NHD{{\mathrm{NHD}}}
\def\Bit{{\mathrm{Bit}}}
\def\floor{{\mathrm{floor}}}
\def\argmin{{\mathrm{argmin}}}

\sysmltitlerunning{Improving Efficiency in Neural Network Accelerator using Operands Hamming Distance Optimization}

\begin{document}

\twocolumn[

\sysmltitle{Improving Efficiency in Neural Network Accelerator using Operands Hamming Distance Optimization}



\sysmlsetsymbol{equal}{*}

\begin{sysmlauthorlist}
\sysmlauthor{Meng Li}{equal,to}
\sysmlauthor{Yilei Li}{equal,to}
\sysmlauthor{Pierce Chuang}{to}
\sysmlauthor{Liangzhen Lai}{to}
\sysmlauthor{Vikas Chandra}{to}
\end{sysmlauthorlist}

\sysmlaffiliation{to}{Facebook, 1 Hacker Way, Menlo Park, CA 94025}

\sysmlcorrespondingauthor{Meng Li}{meng.li@fb.com}
\sysmlcorrespondingauthor{Yilie Li}{yileil@fb.com}


\vskip 0.3in

\begin{abstract}
    Neural network accelerator is a key enabler for the on-device AI inference, for which energy efficiency is an important metric. The datapath energy, including the computation energy and the data movement energy among the arithmetic units, claims a significant part of the total accelerator energy. By revisiting the basic physics of the arithmetic logic circuits, we show that the datapath energy is highly correlated with the bit flips when streaming the input operands into the arithmetic units, defined as the hamming distance of the input operand matrices. Based on the insight, we propose a post-training optimization algorithm and a hamming-distance-aware training algorithm to co-design and co-optimize the accelerator and the network synergistically. The experimental results based on post-layout simulation with MobileNetV2 demonstrate on average 2.85$\times$ datapath energy reduction and up to 8.51$\times$ datapath energy reduction for certain layers.
\end{abstract}
]



\printAffiliationsAndNotice{\sysmlEqualContribution} 

\section{Introduction}
\label{sec:introduction}

Deep neural networks (DNNs) have revolutionized different applications ranging from computer vision to speech and natural language processing \cite{lecun:2015:dl}, and are now widely deployed in data centers \cite{Jouppi:2017, kim:2018:aml_facebook, jongsoo:2018:datacetner} and edge devices \cite{Du:2017, zhang:2017:helloedge, carole:2019:edge}. As modern DNNs usually require significant computation, neural network accelerators are extensively studied in recent years to enable energy-efficient processing \cite{chen:2014:diannao, chen:2016:eyeriss, Sze:2017, Jouppi:2017, hardik:2018:bitfusion}.


Datapath of the neural network accelerator, including the arithmetic compute units and the data bus among the units, lies at the heart of neural network accelerators. It plays an important role in terms of energy consumption of the neural network accelerator. With the trend of aggressive operand quantization ($<$ 8 bit) and near/in-memory computation, the energy consumption of memory accesses in neural network accelerators is greatly reduced. In many state-of-the-art accelerator designs \cite{Andri:2016, Gao:2017, Park:2018}, datapath can consume 40-70\% of the total energy.

Conventionally, the datapath energy consumption in a neural network accelerator can be estimated as $E_{datapath} = \lambda \cdot OPs \cdot Energy/OP$, where $OPs$ denotes the total number of operations of the neural network, $Energy/OP$ is the datapath energy consumption of one operation and $\lambda$ is a correction term that depends on the network parameters and the underlying hardware design,. Previous researches mainly focus on reducing $OPs$, e.g., by optimizing the network topology \cite{forrest:2016:squeezenet, howard:2017, tan:2019:mnasnet} or network pruning \cite{han:2015:deepcompression, he:2017}, and reducing $Energy/OP$, e.g., by network quantization \cite{moons:2016, Park:2018, hardik:2018:bitfusion} or binarization \cite{courbariaux:2016:binarized} etc. In contrast, reducing $\lambda$ receives less attention. Existing works mainly focus on exploiting the sparsity of the network parameters and activations to gate the compute units and skip the unnecessary computations \cite{chen:2016:eyeriss}.


In this work, we explore a new dimension to reduce $\lambda$ and the datapath energy. We show that as most accelerators leverage spatial data reuse \cite{chen:2016:eyeriss} and stream input operands into the compute array, the sequence of the input operands significantly impacts the datapath energy. Specifically, we find that the datapath energy is strongly correlated to the bit flips when streaming the input operands. In this paper, we leverage the concept of hamming distance to formalize the bit flip analysis. A series of post-training and training-aware techniques are proposed to co-design and co-optimize the accelerator and the network to reduce the hamming distance of the input operand sequence. Experimental results based on the post-layout simulation demonstrates on average 3.6$\times$ datapath energy reduction and up to $8.51\times$ energy reduction for certain layers. The proposed techniques are compatible with other optimization knobs, e.g., pruning, quantization, etc. The contributions of the paper can be summarized as follows:


\begin{itemize}
    \item We discover the correlation between the datapath energy and the hamming distance when streaming the input operands and further propose the concept of hamming distance optimization as a new direction of datapath energy optimization;
    \item We propose a post-training optimization algorithm to reduce the hamming distance of the neural network model, which introduces negligible hardware overhead and no impact on the model output;
    \item We propose a hamming-distance-aware training algorithm, which reduces the hamming distance of the neural network model with negligible effect on accuracy;
    \item Experiments based on the post-layout simulation demonstrate promising results (up to 8.51$\times$ datapath energy reduction) by combining the hamming-distance-aware training and the post-processing algorithm.
\end{itemize}
\section{Background: Spatial Accelerators}
\label{sec:background}



Modern NN accelerators usually comprise of the following major components - a two-dimensional arithmetic compute array, a network-on-chip (NoC), control blocks, and an on-chip memory \cite{Sze:2017}. Specifically, the on-chip memory usually consists of several levels of hierarchies, including a global buffer, an inter-unit network to facilitate data pass among the arithmetic units, and register files (RFs) within each arithmetic unit \cite{chen:2016:eyeriss}. The memory access energy to different memory hierarchies can vary significantly.

\begin{figure}[!htb]
    \centering
    \subfloat[]{\includegraphics[width=0.45\linewidth]{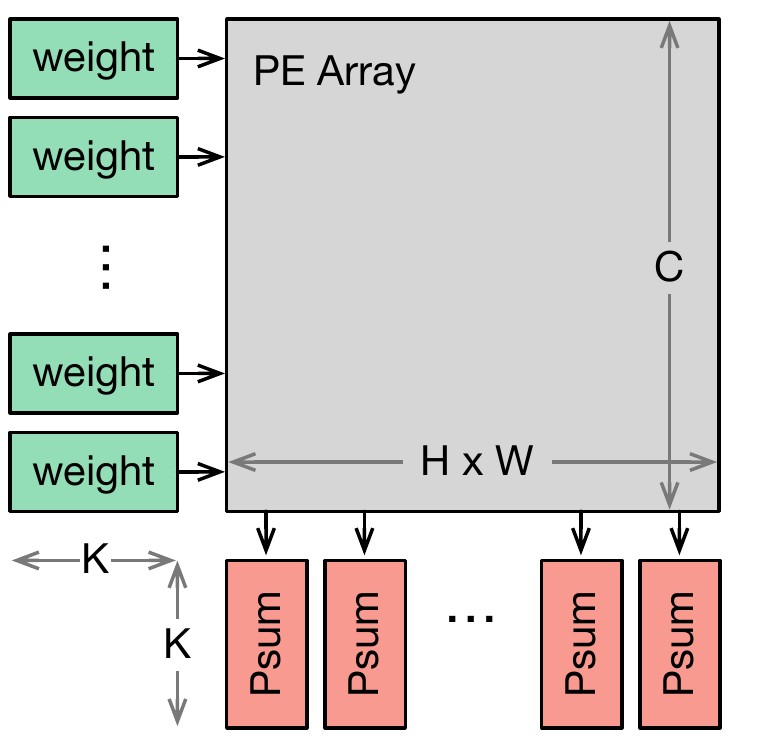}\label{fig:is}}
    \hfill
    \subfloat[]{\includegraphics[width=0.45\linewidth]{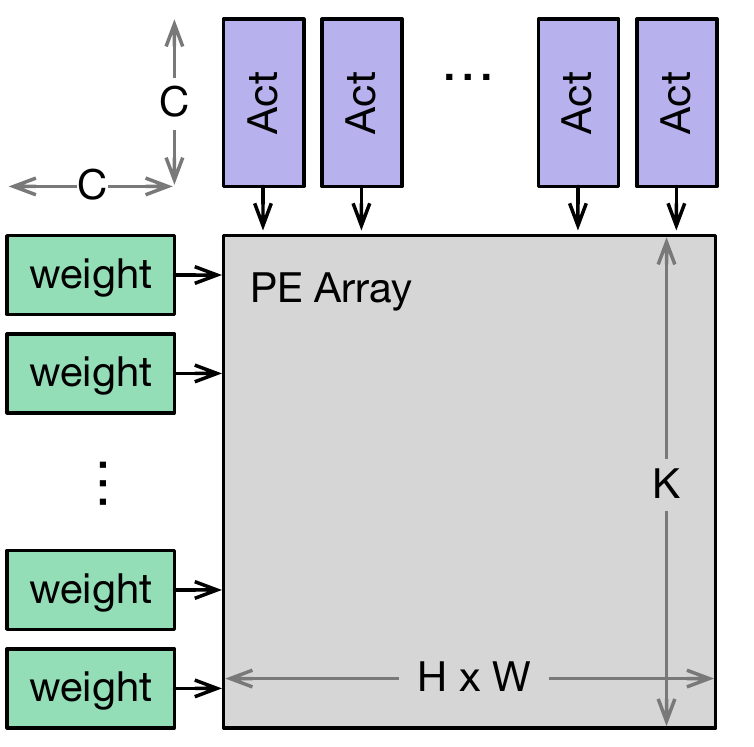}\label{fig:os_hwk}}
    \caption{Different dataflow variants: (a) exploits input stationary dataflows, and (b) leverages the output stationary dataflow. For these dataflow variants, weights are organized in a sequence along either the output channel or the input channel dimension and are sent into the compute array consecutively.}
    \label{fig:dataflow}
\end{figure}

To reduce access to more expensive memory hierarchies, specialized processing dataflows are designed to enable data reuse across different computation units. Representative dataflows include input stationary, output stationary, row stationary, etc \cite{chen:2016:eyeriss, Sze:2017}. The dataflow architecture dictates what data gets read into the memory hierarchy and how data are propagated in the compute array. Figure \ref{fig:dataflow} shows two widely used designs \cite{chen:2018:eyerissv2}. The design in Figure~\subref*{fig:is} leverages the input stationary and relies on unrolling both the input channel dimension ($C$) and input spatial locations ($H \times W$) to map the operations spatially onto the array to exploit the computation parallelism. The weights are streamed into the array and can be reused horizontally with input pixels from different spatial locations, while the partial sums are accumulated spatially across the column. Instead of saving the partial sums directly to the activation SRAM, they are usually stored into an accumulation buffer first to reduce the memory access energy. Until the partial sums are fully reduced, they may go through the nonlinear units and be stored back to the global SRAM. Similarly, the design in Figure~\subref*{fig:os_hwk} leverages the output stationary dataflow and relies on unrolling the output channel dimension ($K$) and output spatial dimensions ($H \times W$) to enable data reuse. In this scheme, the weights are still streamed along the row direction and the input activations are streamed in the orthogonal direction to reuse across different output channels.

\begin{figure}[!htb]
    \centering
    \subfloat[]{\includegraphics[width=\linewidth]{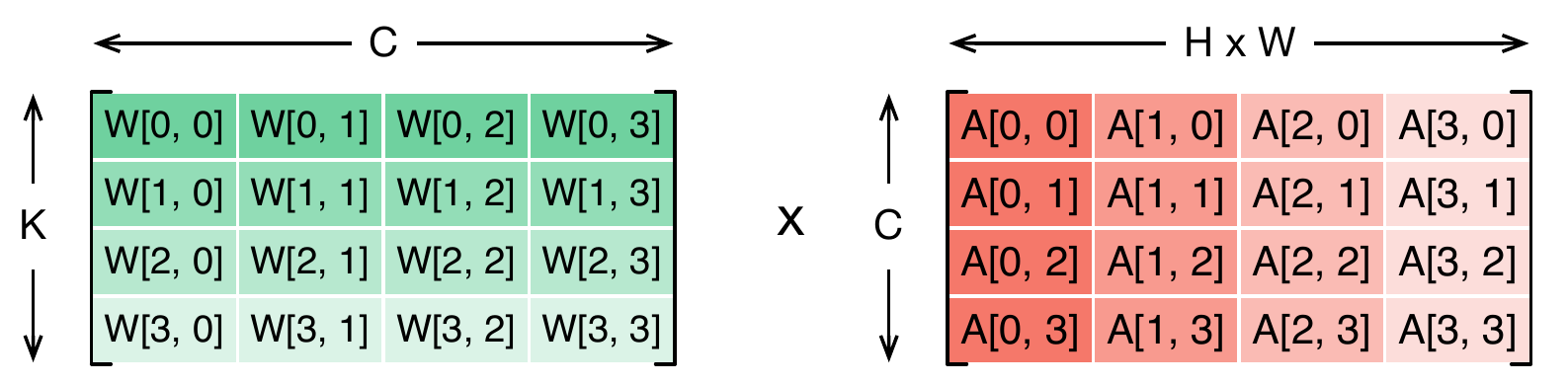}\label{fig:1x1_math}} \\
    \subfloat[]{\includegraphics[width=\linewidth]{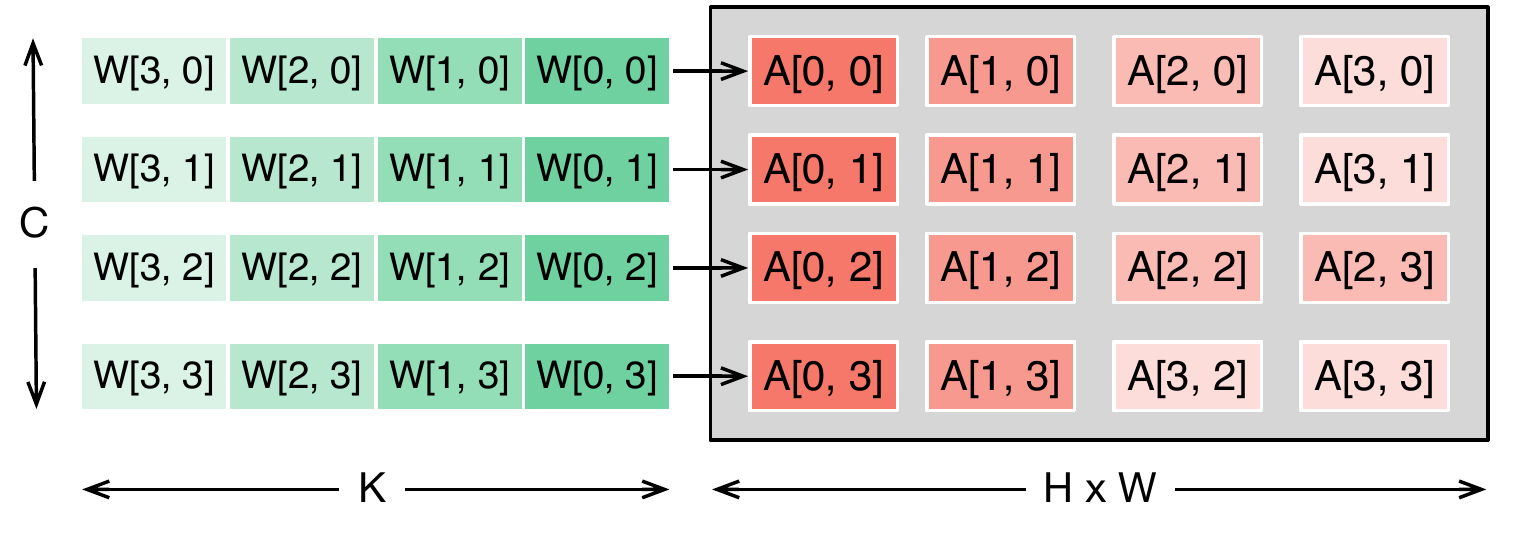}\label{fig:1x1_mapping}}
    \caption{Mapping a 1-by-1 convolution to the input stationary accelerator in Figure~\subref*{fig:is}.}
    \label{fig:1x1_example}
\end{figure}

Popular neural network layers, such as the convolution layer and the fully-connected layer, can be easily mapped to the accelerator. Consider the example of a 1-by-1 convolution in Figure \ref{fig:1x1_example}. To map the computation into the input stationary compute array in Figure~\subref*{fig:is}, the input activations are pre-filled with different input spatial locations unrolled horizontally and different input channels unrolled vertically. The weights are streamed in a sequence into the arithmetic array. For the input stationary dataflow, weights from different input channels are fed spatially into different rows and weights from different output channels are streamed temporally into the same row. 

The energy consumption of the accelerator is composed of the datapath energy (including the arithmetic computation energy and the data propagation energy among compute units), the memory access energy and the control energy. When all the operands can fit into the local SRAM \cite{Park:2018}, the datapath and memory access energy can be computed as
\begin{align*}
    E_{datapath} & = OPs \times Energy/OP \\
    E_{mem}      & = OPs \times (\frac{1}{\tau_{weight}} + \frac{1}{\tau_{input}} + \frac{1}{\tau_{psum}}) \\
    & \quad \times Energy/SRAM Access
\end{align*}
where $\tau_{weight}, \tau_{input}, \tau_{psum}$ denote the reuse factor of the weights, input activation, and the partial sums, respectively. $Energy/OP$ denotes the datapath energy and $Energy/SRAM Access$ denotes the SRAM access energy that includes the SRAM read/write and the data movement energy from SRAM to the compute array.

Assume the ratio between the compute energy, inter-unit propagation energy and the SRAM access energy is 1:2:6 \cite{chen:2016:eyeriss}. For a reasonable design with $\tau_{weight} = 16, \tau_{input} = 16$ and $\tau_{psum} = 16$, the ratio between the datapath energy and the SRAM energy becomes 3: 4 (assuming weights and inputs are 8-bit and partial sums are 32-bit). The datapath energy consumes a significant portion of the total energy and hence it is crucial to reduce the datapath energy.

The datapath energy can be further divided into three parts, including switching energy, glitch energy, and leakage \cite{Rabaey:2008}. Both the switching energy and glitch energy are caused by the circuit nodes switching from 0 to 1 or from 1 to 0, denoted as bit flips. Leakage energy is caused by the small leakage current when the transistors are turned off and its contribution to the datapath energy is usually orders of magnitude smaller than glitch and switching. Hence, we ignore the leakage energy in the paper.

%
%
\section{Motivation: Bit Flips Impact Datapath Energy}
\label{sec:motivation}

As described in Section \ref{sec:background}, while the datapath energy accounts for a significant portion of the total energy, the bit flips inside the datapath are the main culprit. The datapath bit flips are determined by the value and the streaming pattern of the input operands, i.e., weights, input activations, and partial sums. Because the activations and partial sums are input dependent, we focus on analyzing the impact of weight matrices. 

\begin{figure}[!htb]
    \centering
    \includegraphics[width=0.9\linewidth]{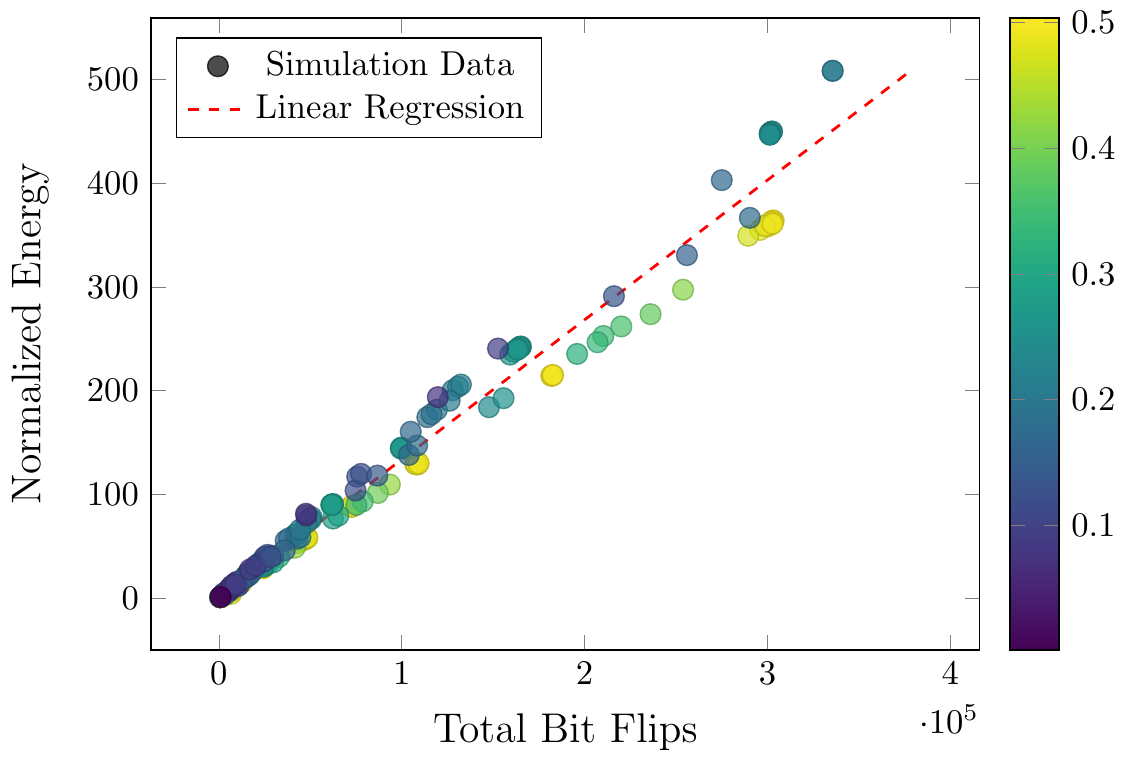}
    \caption{Total bit flips of the weight sequence and the energy consumption demonstrate strong correlation: the colormap represents the average bit flip probability of the input sequence.}
    \label{fig:THD_energy}
\end{figure}

Consider the example of the 2-bit weight matrix $W \in \mathbb{R}^{K \times C}$:
\begin{align*}
    W & = \left.\left[ 
        \vphantom{\begin{array}{c}1\\1\\1\\1\end{array}}
        \smash{
        \underbrace{
        \begin{array}{cccc}
            00 & 00 & 00 & 00 \\
            11 & 11 & 11 & 11 \\
            00 & 00 & 00 & 00 \\
            11 & 11 & 11 & 11
        \end{array} 
        }_{\textcolor{blue}{C}}
        }
        \right]\right\}
        \,\textcolor{blue}{K}.
\end{align*}
\vskip 0.05in
Without loss of generality, we assume an input-stationary compute array as shown in Figure~\subref*{fig:is} and $W$ is streamed into the array following Figure~\subref*{fig:1x1_mapping}. Then, the weight sequence fed into the first row of the compute array is $\{00, 11, 00, 11\}$ and the bit flips of the weight sequence at the compute array input are 6. To confirm the relation between the bit flips of the weight sequence and the datapath energy, we use the weight matrices of MobileNetV2 \cite{sandler:2018:mobilenetv2} trained on Cifar100 dataset as an example and generate random input activations. We evaluate the bit flips of the weight sequence and the datapath energy consumption with post-layout simulation (see Section \ref{sec:results} for detailed experimental setup). As shown in Figure \ref{fig:THD_energy}, the total bit flips of the weight sequence and the energy consumption demonstrate a strong linear relation. Moreover, given a fixed total bit flips, the energy is independent of the length of the weight sequence and the bit flipping probability.

Hence, to minimize the datapath energy, an effective approach is to reduce the bit flips of the weight sequence. We observe that the bit flips can be reduced if the sequence of weight streaming are carefully reordered. Consider $W$ in the example above. If we swap the second row and the third row of the matrix, we have $W^{\prime\prime}$ as below:
\begin{equation*}
    W^{\prime} = 
    \left[\begin{array}{cccc}
    00  & 00  & 00 & 00 \\
    \rowcolor{black!20}
    00  & 00  & 00 & 00 \\
    \rowcolor{black!20}
    11 & 11 & 11 & 11 \\
    11 & 11 & 11 & 11
    \end{array}\right]
\end{equation*}
Now, by streaming $W^{\prime}$ into the compute array, the bit flips can be reduced from $24$ to $8$. \textsl{Note that swapping the rows of the weight matrix is essentially adjusting the order of generating output channels and there is no influence in terms of neural network functionality. As the swapping can be finished via post-processing in model level, no specific hardware support is needed.}

Besides the post-training processing of the weight matrices, another orthogonal approach is to incorporate the bit flips of the weight sequence into the training procedure and reduce the bit flips without sacrificing the model accuracy. Consider $W^{\prime\prime}$ and $v$ below:
\begin{equation*}
    W^{\prime} = 
    \left[\begin{array}{cccc}
    10 & 10 & 10 & 01 \\
    11 & 11 & 11 & 11  \\
    10 & 10 & 10 & 01 \\
    11 & 11 & 11 & 11
    \end{array}\right], 
    v = 
    \left[\begin{array}{c}
    01 \\
    01 \\
    01 \\
    10 
    \end{array}\right]. 
\end{equation*}
While $W v = W^{\prime\prime} v$, the bit flips of the weight sequence for $W^{\prime\prime}$ is $12$. Hence, without impacting the computation results, the bit flips can be reduced by $2\times$. In fact, by further reordering the output channels of $W^{\prime\prime}$, the bit flips can be reduced to $4$. 

In the rest of the section, we will formalize our analysis of the bit flips of the weight sequence and formally describe our post-training and training-aware techniques to reduce the bit flips of the weight sequence.

\section{Methodology: Hamming Distance Optimization}
\label{sec:methodology}


In this section, we will formalize the concept of bit flips and propose both post-training and training-aware techniques to minimize the bit flips of the streaming weights. For convenience, we use the input stationary dataflow (e.g., Figure \subref*{fig:is}) as an example throughout the analysis but the definition, analysis, and conclusion can be easily applied to other dataflow schemes once the weights are streamed into the compute array. The notations used in this paper are summarized in Table \ref{tab:notation}.

\begin{table}[!htb]
\caption{Notations used in the paper.}
\label{tab:notation}
\begin{tabular}{c|c}
\hline \hline
$N, H, W, K$   & Output batch, height, width, channel \\ \hline
$C, F_x, F_y$  & Input channel, filter height, width\\ \hline
$W$          & Model weight matrix \\ \hline
$B$          & Bit width of model weights \\ \hline
$S$          & Sequence of output channels \\ \hline
$T$          & Cluster of input channels \\ \hline
\end{tabular}
\end{table}

\subsection{Problem Formulation}
\label{subsec:formulation}

In coding theory, the bit difference between two binary strings are formally defined as the hamming distance \cite{Hamming:1950}. Accordingly, we define the hamming distance between two $B$-bit numbers $a$ and $b$ as
\begin{equation*}
    \HD(a, b) = \sum_{i=1}^{B} \Bit_{i}(a) \oplus \Bit_{i}(b),
\end{equation*}
where $\oplus$ denotes the XOR operation and $\Bit_i(\cdot)$ is the function that extracts the $i$-th bit of the number. 

\begin{figure}[!tb]
    \centering
    \includegraphics[width=0.7\linewidth]{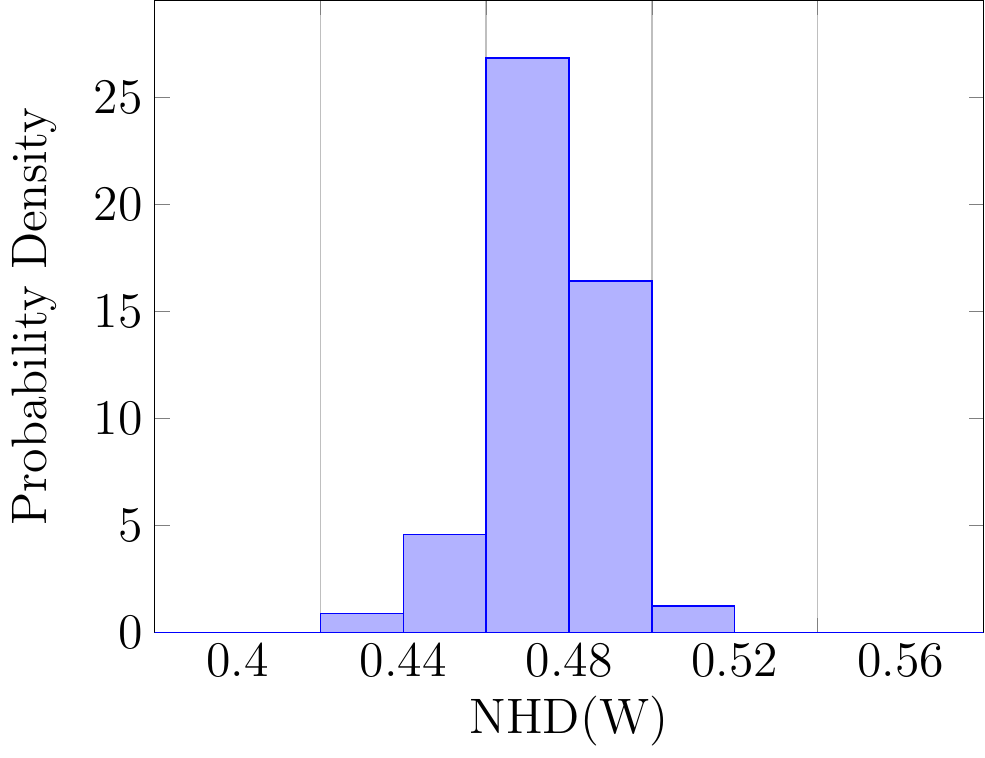}
    \caption{$\NHD(W)$ distribution for different layers in MobileNetV2 and ResNet26.}
    \label{fig:nhd}
\end{figure}

Consider a weight matrix $W \in \mathbb{R}^{K \times C}$ \footnote{We assume $F_x = F_y = 1$ for the weight matrix in this case, but the definition and analysis can be easily extended to cases where $F_x$ and $F_y$ are larger than 1.}. As the input stationary dataflow unrolls the input channel dimension ($C$) along the compute array column direction and stream the weights along different output channels ($K$) in temporal sequence to the array, we define the hamming distance of streaming $W$ as
\begin{align*}
    \HD(W) & = \sum_{j=1}^{K-1} \HD(W[j, :], W[j+1, :]) \\
           & = \sum_{j=1}^{K-1} \sum_{i=1}^{C} \HD(W[j, i], W[j+1, i])
\end{align*}
We also define the normalized hamming distance (NHD) of streaming $W$ as 
\begin{align*}
    \NHD(W) = \frac{\HD(W)}{C \times (K-1) \times B}.
\end{align*}
Hence, $\HD(W)$ captures the total bit flips of streaming $W$ and $\NHD(W)$ represents the bit flip probability. We show $\NHD(W)$ for different layers of the MobileNetV2 and ResNet26 trained on Cifar100 in Figure \ref{fig:nhd} and as we can see, $\NHD(W)$ is close to 0.5 for all the layers. In the following sections, we will propose techniques to minimize $\HD(W)$ and $\NHD(W)$ to reduce the bit flips and the datapath energy.

\subsection{Output Channel Reordering}
\label{subsec:output_reordering}

Inspired by the example in Section \ref{sec:motivation}, a straightforward technique to minimize $\HD(W)$ is to reorder the sequence of $W$ streaming into the compute array. Let $S$ denote the sequence of output channels to stream $W$ into the array and $\HD_{S}(W)$ denote the hamming distance of streaming $W$ following $S$, then, we have
\begin{equation*}
    \HD_{S}(W) = \sum_{j=1}^{K-1} \sum_{i=1}^{C} \HD(W[S[j], i], W[S[j+1], i]).
\end{equation*}
The output channel reordering problem is defined as follows.

\begin{problem}{(Output Channel Reordering)}
Given a weight matrix $W \in \mathbb{R}^{K \times C}$, find $S^*$ such that $\HD_{S^*}(W)$ is minimized, i.e.,
\begin{equation*}
    S^* = \argmin_{S} \HD_{S}(W).
\end{equation*}
\end{problem}

As $S$ is a reordering of the output channels which consists of each output channel exactly once, we map the reordering problem to a Traveling Salesman Problem (TSP) \cite{Miller:1960}. Specifically, each output channel $i$ corresponds to one location to visit, and the hamming distance between two output channels $i$ and $j$, i.e., $\HD(W[i, :], W[j, :])$, corresponds to the distance between two locations. Hence, minimizing $\HD_{S}(W)$ is equivalent to searching for the shortest path to visit all the locations. Hence the complexity of solving the output channel reordering problem scales exponentially, which quickly becomes intractable for moderate size problems.

To efficiently solve the reordering problem, we propose a greedy search algorithm as described in Algorithm \ref{alg:k_reordering}. The algorithm first initializes the sequence $S$ by assigning the first output channel to the starting position of $S$. After that, the output channel that has the smallest hamming distance compared with the previous channel in $S$ is added to $S$. The complexity of the algorithm scales quadratically with the number of output channels, which is very efficient in practice. 

\begin{algorithm}[!htb]
\SetAlgoLined
\textbf{Input:} weight matrix $W$ \;

\textbf{Output:} optimal sequence $S$ that minimizes $\HD(W)$ \;

$S \leftarrow $ \textsc{Initialize()} \;

\For{$i = 2: K$}{
    $j \leftarrow \argmin_j \HD(W[S[i-1], :], W[j, :])$ \; \\
    $S[i] \leftarrow j$ \;
}
\caption{Greedy Output Channel Reordering.}
\label{alg:k_reordering}
\end{algorithm}

\subsection{Input Channel Segmentation and Clustering}
\label{subsec:c_clustering}

While the output channel reordering can help reduce the hamming distance of streaming $W$, the effectiveness is impacted by the number of input channels $C$. We use MobileNetV2 on Cifar100 dataset \cite{krizhevsky:2009:cifar} as an example and evaluate the hamming distance reduction for different layers. As shown in Table \ref{tab:hd_impact_ck}, with the increase of $C$, the hamming distance reduction slows down significantly.

\begin{table}[!tb]
\caption{Hamming Distance Reduction with Various $C$ and $K$.}
\label{tab:hd_impact_ck}
\vskip 0.15in
\begin{center}
\begin{sc}
\begin{tabular}{l|cccc}
\toprule
layer & $C$ & $K$ & HD Reduction \\        
\midrule
layer 7      & 192    & 32      & 1.53$\times$    \\        
layer 15     & 384    & 64      & 1.33$\times$    \\        
layer 21     & 576    & 96      & 1.27$\times$    \\        
layer 27     & 960    & 160     & 1.18$\times$    \\        
layer 33     & 1280   & 320     & 1.21$\times$    \\        
\bottomrule
\end{tabular}
\end{sc}
\end{center}
\vskip -0.1in
\end{table}

One straightforward method to improve the effectiveness of the output channel reordering is to segment the weight matrix $W$ along the input channel direction into several small sub-matrices. For different sub-matrices, we can use Algorithm \ref{alg:k_reordering} to search for the optimal output channel order to reduce the hamming distance. We denote this method as the segment-then-reorder approach. It should be noted that as the output channel sequence changes, specific hardware support in the accumulator is required to make sure the partial sums corresponding to the same output channel are correctly accumulated. We will detail the hardware support in Section \ref{sec:hw_sup}, which introduces negligible overhead to the accumulator. With the segment-then-reorder approach, the hamming distance can be further reduced by 1.5-2.5$\times$ compared with the direct output channel reordering (see Section \ref{sec:results}).



As expected, the smaller each input channel group is, the better the hamming distance reduction can be achieved. Hence, the segment-then-reorder algorithm would favor the compute array with a skewed aspect ratio, i.e., more columns and fewer rows. However, the aspect ratio of the compute array also impacts the reuse of different operands \cite{chen:2016:eyeriss} and utilization. For example, with more number of columns in the input stationary array, it takes more pixels in the spatial plane to fill the whole array and thus, leads to under utilization for small input activation sizes. While this is not a problem for small-scale arrays with a small number of compute units, it may induce utilization issue for large arrays.

\begin{figure*}[!htb]
    \centering
    \subfloat[]{\includegraphics[height=0.29\linewidth]{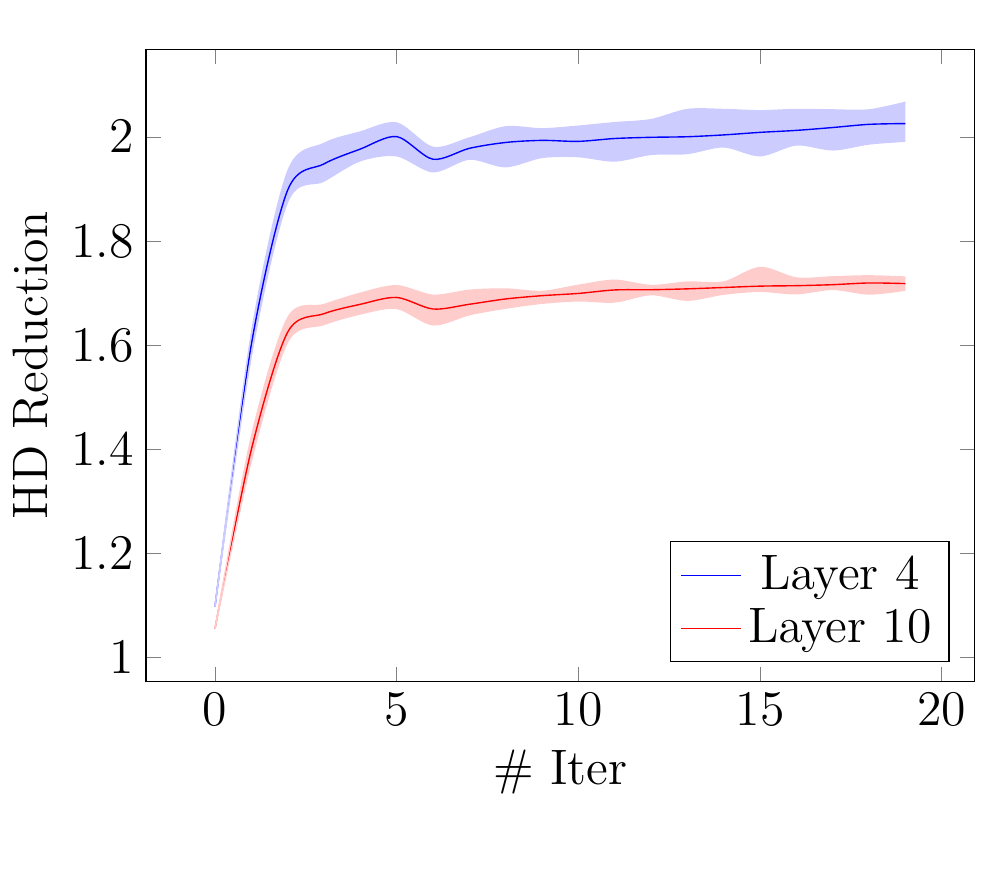}\label{fig:convergence}}
    \hfill
    \subfloat[]{\includegraphics[height=0.29\linewidth]{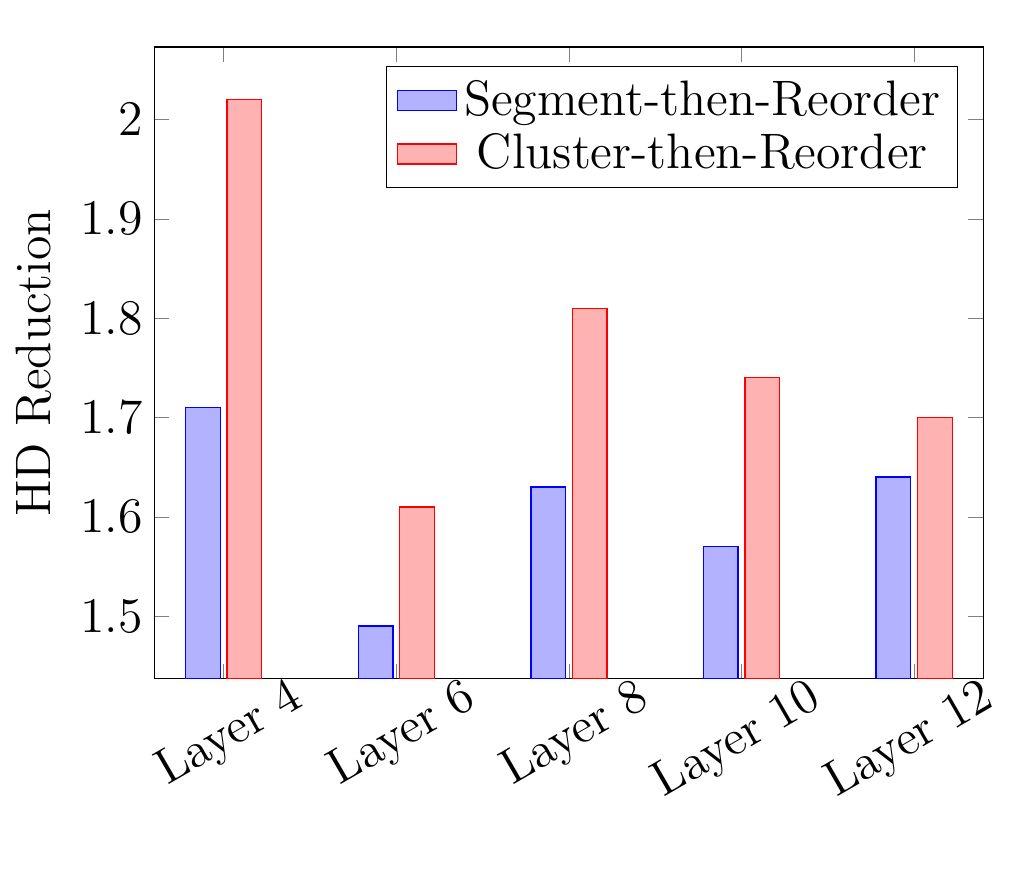}\label{fig:effective}}
    \hfill
    \subfloat[]{\includegraphics[height=0.29\linewidth]{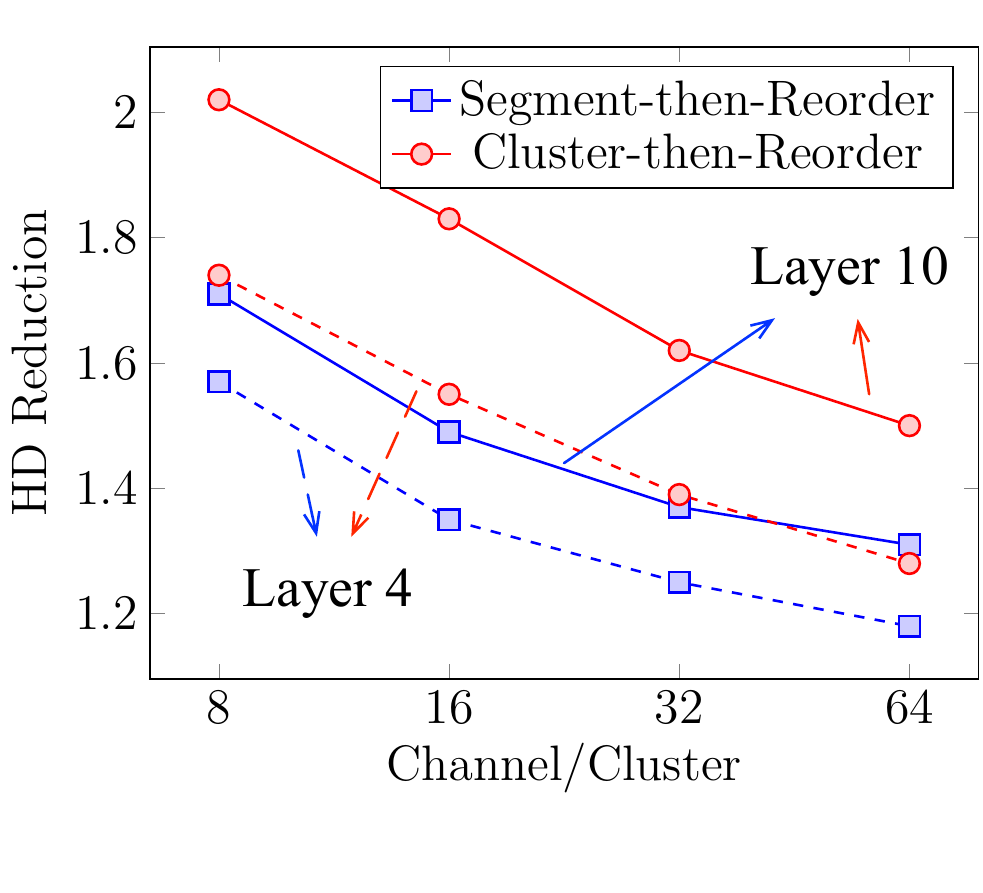}\label{fig:cluster_num}}
    \caption{Performance of the cluster-then-reorder algorithm: (a) convergence plot; and normalized hamming distance comparison with the segment-then-reorder algorithm for (b) different layers (channels per cluster is 8) and (c) different channels per cluster.}
    \label{fig:cluster_comparison}
\end{figure*}

To further improve the effectiveness when the input channel per segment is large, we propose to cluster the input channels first before segmenting the weight matrix. Then, the output channels are reordered for each cluster separately. We denote this approach as cluster-then-reorder.


Consider the example of the following weight matrix:
\begin{equation*}
    W = \left[\begin{array}{>{\columncolor{black!10}}cc>{\columncolor{black!10}}cc>{\columncolor{black!10}}cc>{\columncolor{black!10}}cc}
    00 & 11 & 00 & 11 & 01 & 10 & 01 & 10 \\
    11 & 11 & 00 & 00 & 10 & 10 & 01 & 01 \\
    11 & 00 & 00 & 11 & 10 & 01 & 10 & 10 \\
    11 & 11 & 11 & 11 & 10 & 10 & 10 & 10
    \end{array}\right]
\end{equation*}
Assume the compute array has 4 rows and only allows for streaming 4 input channels simultaneously. Instead of directly segmenting $W$, we can first cluster the input channels into 2 groups, i.e., $\{0, 2, 4, 6\}$ and $\{1, 3, 5, 7\}$, and then segment $W$ into $W^\prime$ and $W^{\prime\prime}$ as below. Compared to the segment-then-reorder approach, the hamming distance can be reduced from 22 to 16. Note that the clustering of the input channels does not impact the output.

\begin{equation*}
    W^\prime = \left[\begin{array}{>{\columncolor{black!10}}c>{\columncolor{black!10}}c>{\columncolor{black!10}}c>{\columncolor{black!10}}c}
    00 & 00 & 01 & 01 \\
    11 & 00 & 10 & 01 \\
    11 & 00 & 10 & 10 \\
    11 & 11 & 10 & 10 
    \end{array}\right],
    W^{\prime\prime} = \begin{bmatrix}
    11 & 11 & 10 & 10 \\
    11 & 00 & 10 & 01 \\
    00 & 11 & 01 & 10 \\
    11 & 11 & 10 & 10 
    \end{bmatrix}.
\end{equation*}

Let $\{T_1, \ldots, T_t\}$ denote the $t$ clusters of the input channel. The input channel clustering problem is then defined as follows.

\begin{problem}(Input Channel Clustering)
Given a weight matrix $W \in \mathbb{R}^{K \times C}$, find $t$ clusters $T_1, \ldots, T_t$ such that the total hamming distance of streaming each sub-matrix $W_{T_i}$ is minimized, i.e., 
\begin{align*}
    \min_{T_1, \ldots, T_t} & \quad \sum_{i=1}^{t} \HD_{S_i^*}(W_{T_i}) \\
    \mathrm{s.t.} & \quad S_i^* = \argmin_{S} \HD_{S}(W_{T_i}) \\
                  & \quad T_i \cap T_j = \emptyset \quad \forall i \neq j \\
                  & \quad \cup_{i=1}^{t} T_i = \{1, \ldots, C\}
\end{align*}
\end{problem}
This is a nested optimization problem which is computationally expensive to solve optimally even if the inner optimization loop can be solved with the proposed Algorithm \ref{alg:k_reordering}. Hence, we propose a greedy iterative method to solve the nested optimization problem. As shown in Algorithm \ref{alg: c_clustering}, in the initialization process, $t$ input channels are randomly selected and $\{S_1^{(0)}, \ldots, S_t^{(0)}\}$ are initialized to minimize the total hamming distance for each input channel. The algorithm alternates between the assignment step and the update step for $N$ total iterations. In the assignment step, for each input channel $i$, we evaluate its hamming distance following the optimal sequence of each cluster, i.e., $S^{(n)}_1, \ldots, S^{(n)}_t$. The input channel is then added to the cluster with the smallest hamming distance. In the update step, we re-compute the optimal sequence for each cluster of the input channels.

\begin{algorithm}[!tb]
\SetAlgoLined
\textbf{Input:} weight matrix $W \in \mathbb{R}^{K \times C}$, number of iterations $N$, number of clusters $t$ \;

\textbf{Output:} cluster of input channels $\{T_1^*, \ldots, T_t^*\}$ and the optimal sequence of output channels $\{S_1^*, \ldots, S_t^*\}$ \;

$\{S_1^{(0)}, \ldots, S_t^{(0)}\}, n \leftarrow$ \textsc{Random\_Initialize}(), 0\;

\While{$n \leq N$}{
    \textcolor{blue}{\tcp{Assignment step}}
    \For{$i = 1: C$}{
        $l \leftarrow \argmin_{k} \HD_{S_k^{(n)}}(W[:, i])$\;
        
        $T_l^{(n)} \leftarrow T_l^{(n)} \bigcup \{i\}$\;
    }
    \textcolor{blue}{\tcp{Update step}}
    \For{$i = 1: t$}{
        $S_i^{(n + 1)} \leftarrow \argmin_{S} \HD_{S}(W_{T_i})$\;
    }
    $n += 1$\;
}

$\{S_1^*, \ldots, S_t^*\} \leftarrow \{S_1^{(N)}, \ldots, S_t^{(N)}\}$ \;

$\{T_1^*, \ldots, T_t^*\} \leftarrow \{T_1^{(N)}, \ldots, T_t^{(N)}\}$ \;

\caption{Cluster-then-reorder Algorithm}
\label{alg: c_clustering}
\end{algorithm}

The convergence of the proposed clustering algorithm can be guaranteed if the inner loop optimization, i.e., the output channel reordering problem, can be optimally solved. This is because the objective function of the clustering problem is always bounded and it is guaranteed to be reduced in the assignment and update step in each iteration. In practice, we use the greedy algorithm to solve the update step as described in Section \ref{subsec:output_reordering}. We find the cluster-then-reorder algorithm converges very well and continuously out-perform the segment-then-reorder algorithm. 


We use the layers of the MobileNetV2 \cite{sandler:2018:mobilenetv2} on Cifar100 dataset as an example and run the clustering algorithm 20 times with random initialization. The convergence plot is shown in Figure~\subref*{fig:convergence}. The normalized hamming distance is computed as the hamming distance of different algorithms normalized by the hamming distance without output channel reordering. As we can see, the clustering algorithm converges within 15 iterations and the run-to-run variation of the clustering algorithm is very small. We also compare the cluster-then-reorder algorithm with the segment-then-reorder algorithm for different layers and different numbers of channels per cluster. As shown in Figure~\subref*{fig:effective} and \subref*{fig:cluster_num}, the cluster-then-reorder algorithm can out-perform the baseline algorithm by up to 1.21$\times$.


The proposed algorithm is very efficient since the complexity of the update step scales $O(K^2C)$ and the complexity of the assignment step scales $O(CKt)$ with the number of input channels $C$, output channels $K$, and the number of clusters $t$.


\subsection{Hamming Distance-Aware Training}
\label{subsec: thd_training}

While the techniques proposed above focus on post-training optimization, we also propose a hamming distance-aware training procedure to further reduce the hamming distance of streaming $W$. The basic idea is to incorporate the hamming distance loss into the loss function and explicitly encourage the reduction of hamming distance as shown below:
\begin{align*}
    \gL = \gL_{CE} + \lambda \gL_{HD}(W),
\end{align*}
where $\gL_{CE}$ represents the original cross-entropy loss. $\lambda$ is used to explicitly control the trade-off between the accuracy and the hamming distance reduction.

However, there are two main problems with $\gL_{HD}(W)$. Firstly, to compute $\gL_{HD}$, $\Bit(\cdot)$ is needed. Consider a integer $x$, to get the $b$-th bit, we have
\begin{equation*}
    \Bit_b(x) = \frac{1}{2^b} \floor(x - 2^{b+1} \floor(\frac{x}{2^{b+1}})).
\end{equation*}
Because $\floor(\cdot)$ is not differentiable, $\gL_{HD}$ is not differentiable as well.

Previously, straight-through-estimator (STE) has been proposed to approximate the gradients for $\floor(\cdot)$ \cite{bengio:2013}. However, directly applying STE leads to 
\begin{align*}
    \frac{\partial \Bit_b}{\partial x} = 0, \quad \forall b \neq B - 1.
\end{align*}
This indicates that only the most significant bit of the weight parameters can be regularized. Hence, we propose an iterative freeze-and-regularize procedure. In the network training process, we first add regularization to the most significant bit and after several epochs, we freeze the most significant bit and after that regularize the second most significant bit. The iterative process continues until we fix all the bits of the weights. 

\begin{figure}[!tb]
    \centering
    \includegraphics[width=0.8\linewidth]{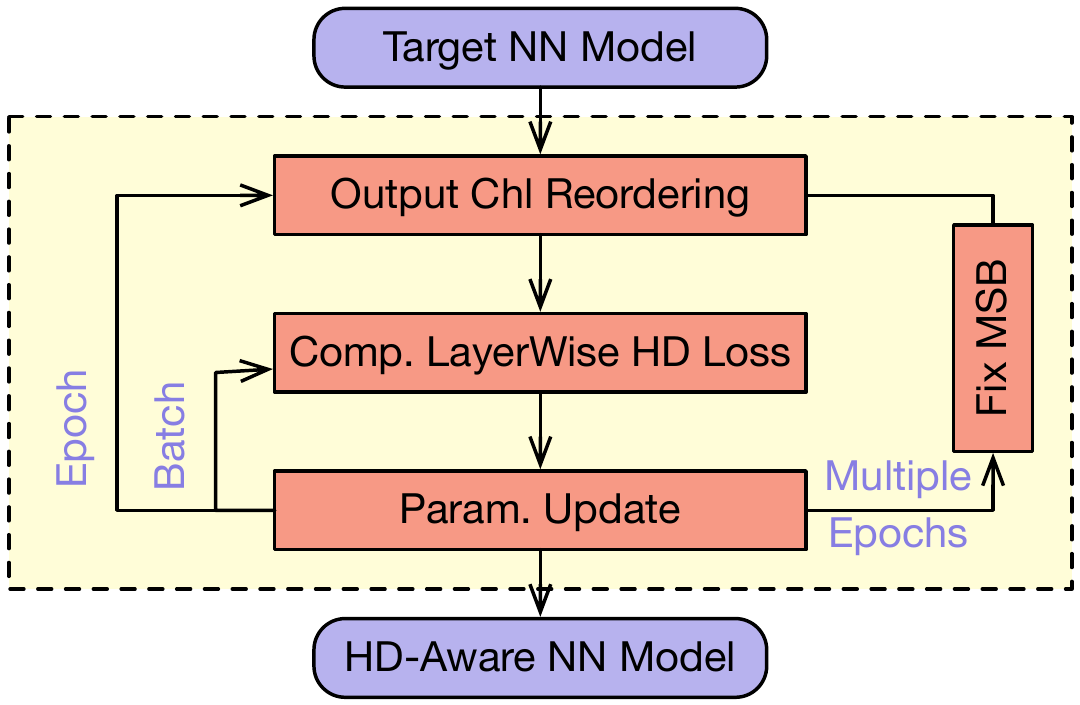}
    \caption{Hamming distance-aware training procedure.}
    \label{fig:training_aware}
\end{figure}

The second problem with $\gL_{HD}$ is that to compute $\gL_{HD}$, the input channel clusters and output channel orders are needed. As the weight matrices get updated during training, both the optimal input channel clusters and the optimal output channel order can change. Hence, after each epoch of training, we leverage the cluster-then-reorder algorithm to cluster the input channels and reorder the output channels. The final training procedure is shown in Figure \ref{fig:training_aware}.

\section{Hardware Support}
\label{sec:hw_sup}

In this section, we discuss the necessary hardware support for the proposed algorithms, including direct greedy reorder, segment-then-reorder, and cluster-then-reorder algorithms.

\begin{figure}[!tb]
    \centering
    \subfloat[]{\includegraphics[width=0.35\linewidth]{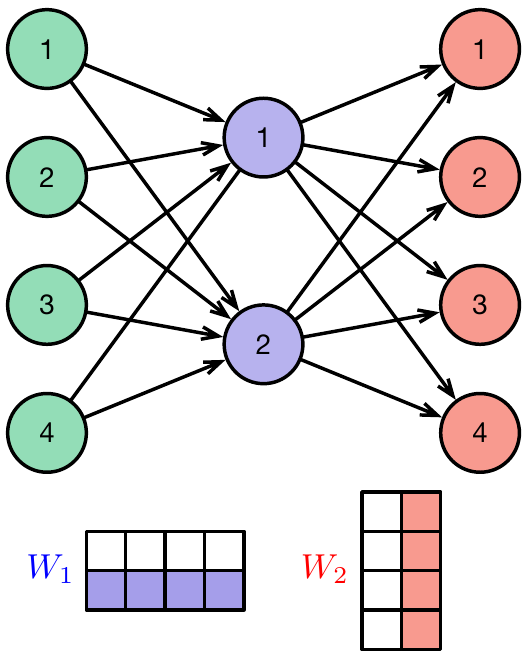}\label{fig:k_reorder_hw_1}}
    \hspace{30pt}
    \subfloat[]{\includegraphics[width=0.35\linewidth]{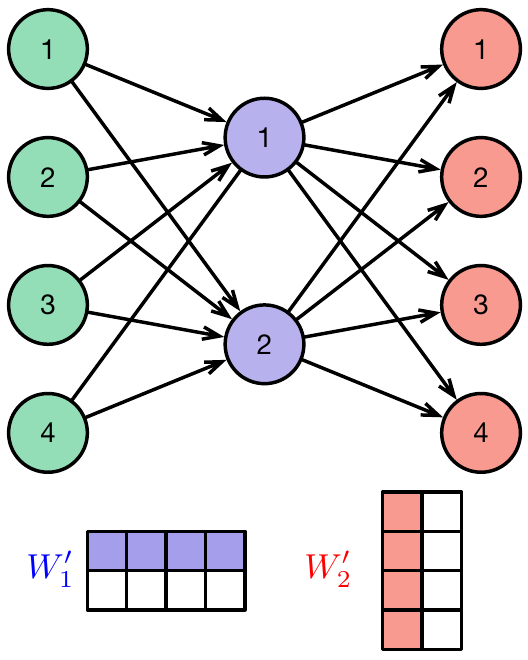}\label{fig:k_reorder_hw_2}}
    \caption{Post-training processing to reorder the output channel.}
    \label{fig:k_reorder_hw}
\end{figure}

The direct reorder algorithm only switches the sequence for the output channel generation. No extra hardware support is needed for the direct reorder algorithm. Instead, a post-training processing of the model to re-arrange the weight matrices is sufficient. Consider the example in Figure \ref{fig:k_reorder_hw}. To switch the output channels of the first layer, both the rows of the weight matrix in the first layer, i.e., $W_1$, and the columns of the weight matrix in the second layer, i.e., $W_2$, need to be switched accordingly.

\begin{figure}[!tb]
    \centering
    \subfloat[]{\includegraphics[width=0.80\linewidth]{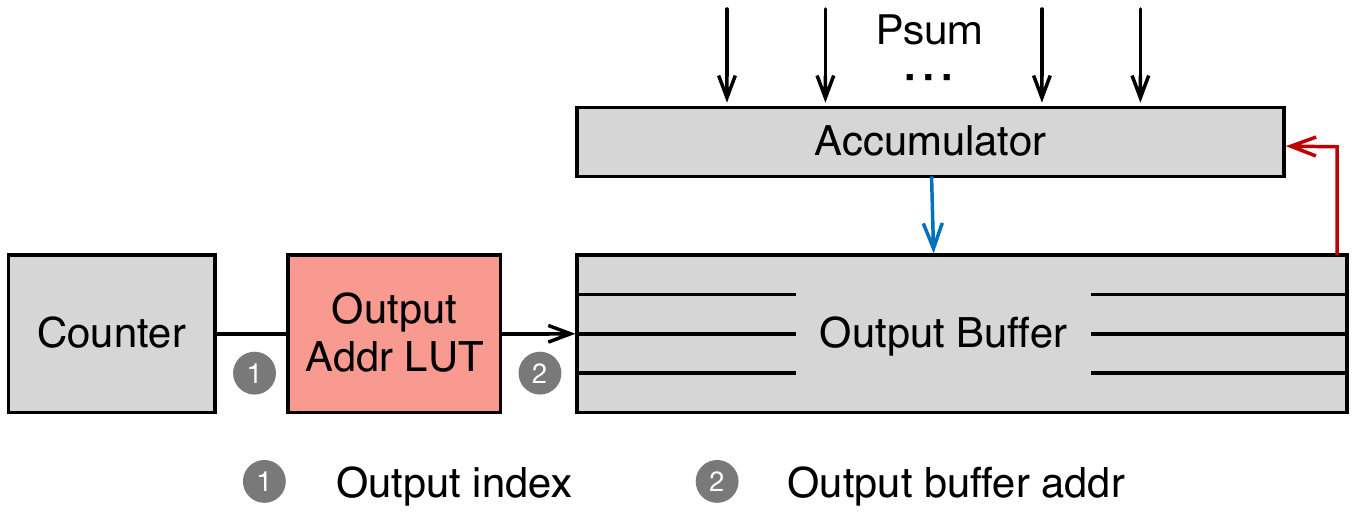}\label{fig:output_lut}} \\
    \subfloat[]{\includegraphics[width=\linewidth]{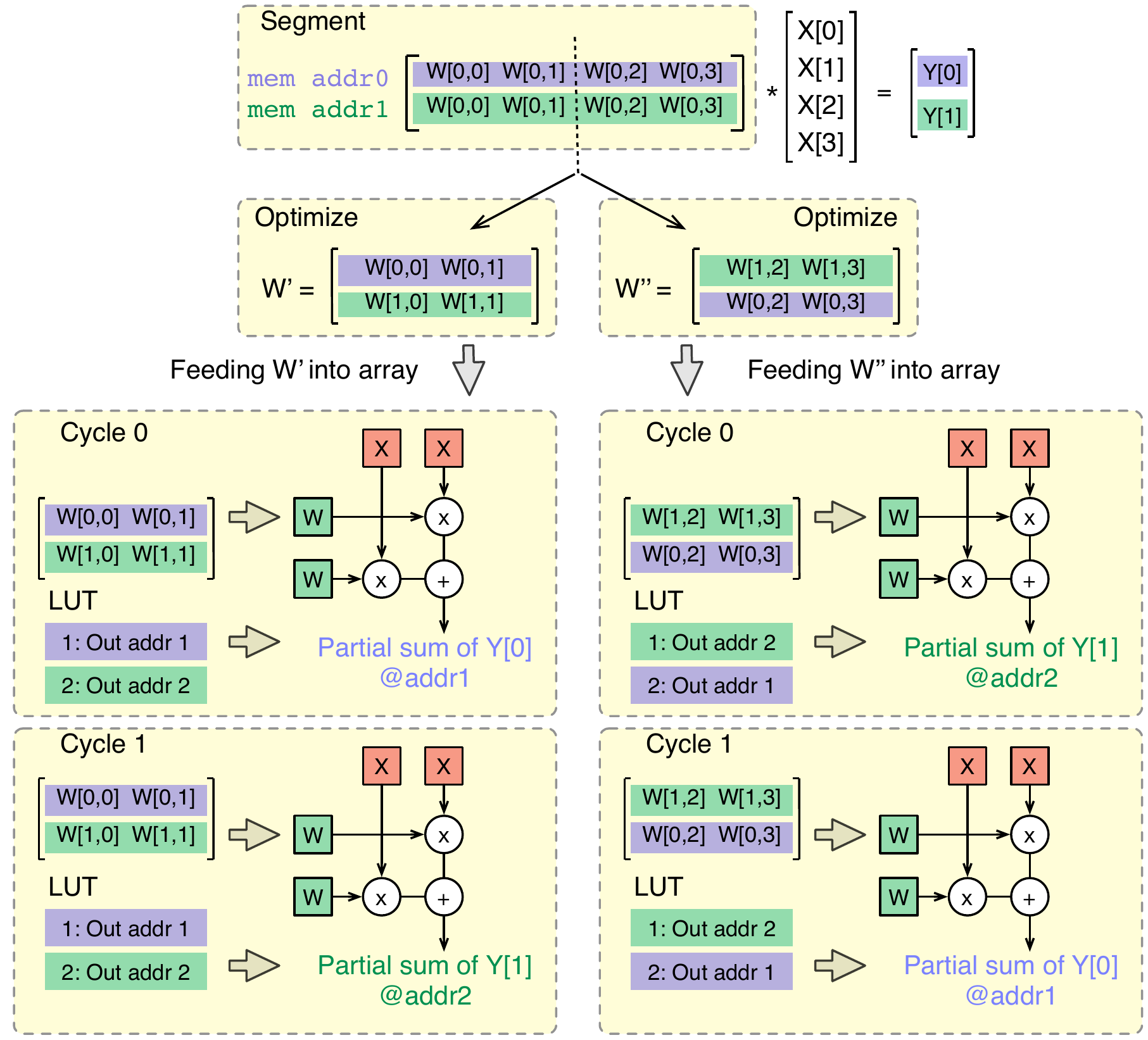}\label{fig:output_lut_exg}}
    \caption{(a) Hardware support for the segment-then-reorder algorithm and (b) an example with 2 segments.}
    \label{fig:segment_the_reorder}
    \vskip -0.2in
\end{figure}

The segment-then-reorder algorithm divides the input channels into segments and reorders the output channels for each segment separately. Hence, the same row in different segmented weight sub-matrices may correspond to the partial sum of different output channels. To guarantee correct reduction of the partial sums, we add an output address lookup table (LUT) to translate the index of the counter in the accumulator to the actual address for accumulation as shown in Figure~\subref*{fig:output_lut}. We also show in Figure~\subref*{fig:output_lut_exg} an example on how to use the address LUT to guide the accumulation. As can be seen, by modifying the LUT entry corresponding to different counter indices, the partial sums are correctly accumulated.

\begin{figure*}[!tb]
    \centering
    \includegraphics[width=\linewidth]{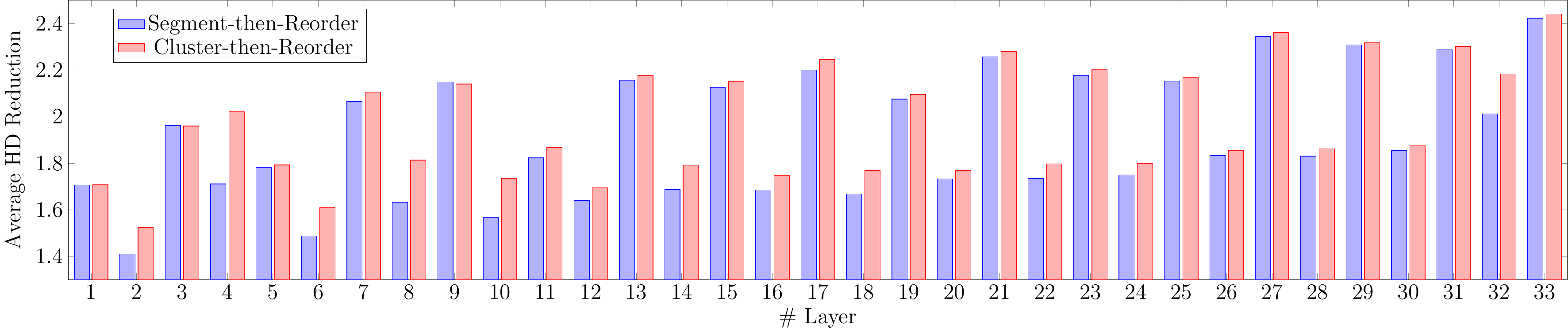}
    \caption{Hamming distance reduction comparison for the segment-then-reorder and the cluster-then-reorder algorithm on MobileNetV2.}
    \label{fig:per_layer_comp}
\end{figure*}

If we assume the output buffer depth to be $D$, the LUT needs to have at least $D$ entries and each entry needs to have $\log_2D$ bits. For a reasonable output buffer depth, e.g., 1024, the LUT SRAM size is less than 2 KB, which is very small and thus has negligible energy and area overhead.

Compared to the segment-then-reorder algorithm, the cluster-then-reorder algorithm also changes the order of the input channels, i.e., the columns of the weight matrices. While the clustering does not impact the correctness of the outputs, it may impact the memory fetching of the input activations. We leverage the output address LUT to swap the activations to avoid any complication or modification to the input fetching logic. For example, let's assume the required input channel sequence for the current layer to be $\{1, 3, 4, 2\}$. When executing the previous layer, the output address LUT can simply be set to $\{1: 1, 2: 3, 3: 4, 4: 2\}$ to reorder the sequence of channel generation.

\section{Experimental Results}
\label{sec:results}

\subsection{Experimental Setup}


In this section, we report on our experiments to demonstrate the effectiveness of the proposed hamming distance reduction techniques. We use MobileNetV2 \cite{sandler:2018:mobilenetv2} and ResNet26 \cite{he:2016:resnet} trained on the Cifar10 and Cifar100 dataset for the evaluation. The 1-by-1 convolution layers in MobileNetV2 and the 3-by-3 convolution layers in ResNet26 are picked \footnote{3-by-3 depthwise separable convolutions are not considered as they are usually hard to map on the systolic arrays and they only consume a very small part of the total energy.}. The layer shapes are shown in \ref{sec:layer_shapes_app}. To evaluate the energy consumption, we use simulation on a post-layout extracted netlist. We designed an input-stationary systolic array with 8 rows and 8 columns. Each PE in the array can support the multiplication and accumulation of 8-bit activations and 4-bit weights. The array is synthesized and placed and routed using a commercial technology library and the energy consumption is evaluated in a typical process corner. The leakage energy is ignored in the evaluation as it is more than two orders of magnitude less than dynamic energy.

\begin{table*}[!tb]
\def\arraystretch{1.1}
\caption{Training-aware hamming distance optimization for MobileNetV2 on Cifar10 and Cifar100 dataset.}
\label{tab:training_aware}
\resizebox{\linewidth}{!}{
\begin{sc}
\begin{tabular}{c|c|cc|cc|cc}
\toprule
Dataset                   & $\lambda$ & Top-1 Acc & Top-5 Acc & \begin{tabular}[c]{@{}c@{}}Best-Layer HD\\ Reduction\end{tabular} & \begin{tabular}[c]{@{}c@{}}Average HD\\ Reduction\end{tabular} & \begin{tabular}[c]{@{}c@{}}Best-Layer Energy\\ Reduction\end{tabular} & \begin{tabular}[c]{@{}c@{}}Average Energy\\ Reduction\end{tabular} \\
\midrule
\multirow{2}{*}{Cifar10}  & $0.0$               & 94.38  & 99.82     & 1.0$\times$   & 1.0$\times$  & 1.0$\times$     & 1.0$\times$         \\
                          & $1 \times 10^{-4}$  & 94.22  & 99.00     & 41.6$\times$  & 7.55$\times$ & 18.6$\times$    & 6.63$\times$        \\
\midrule
\multirow{5}{*}{Cifar100} & $0.0$               & 78.21  & 94.53     & 1.0$\times$   & 1.0$\times$  & 1.0$\times$     & 1.0$\times$         \\
                          & $1 \times 10^{-5}$  & 77.98  & 94.20     & 2.31$\times$  & 1.24$\times$ & 2.17$\times$    & 1.26$\times$        \\
                          & $3 \times 10^{-5}$  & 77.47  & 94.07     & 3.19$\times$  & 1.50$\times$ & 2.86$\times$    & 1.47$\times$        \\
                          & $5 \times 10^{-5}$  & 77.29  & 94.24     & 4.54$\times$  & 1.76$\times$ & 3.88$\times$    & 1.67$\times$        \\
\rowcolor{black!10}
                          & $7 \times 10^{-5}$  & 77.62  & 94.26     & 5.95$\times$  & 2.00$\times$ & 4.92$\times$    & 1.86$\times$        \\
\bottomrule
\end{tabular}
\end{sc}
}
\end{table*}


\subsection{Post-Training Hamming Distance Optimization}

\begin{figure}[!tb]
    \centering
    \subfloat[]{\includegraphics[width=0.9\linewidth]{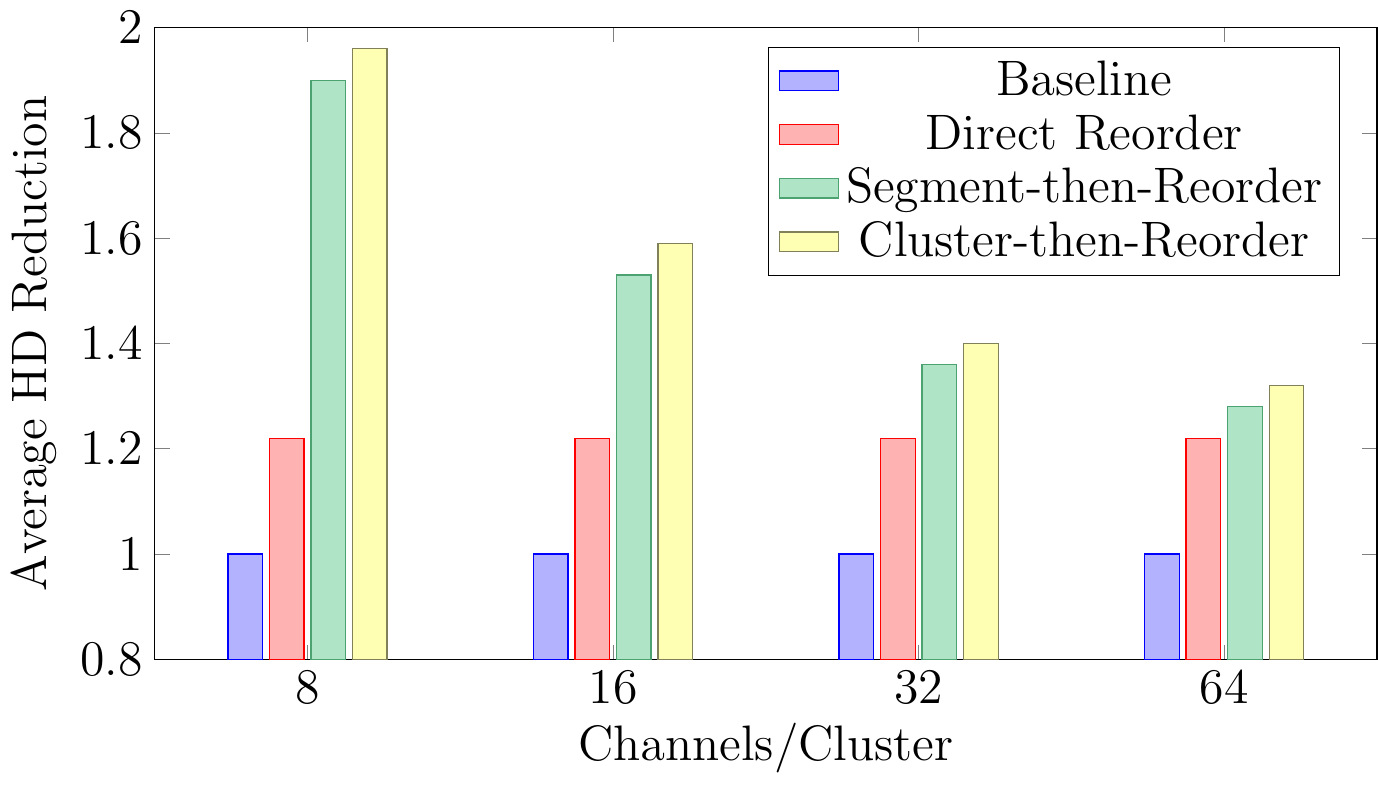}\label{fig:post_train_mobilenet}} \\
    \subfloat[]{\includegraphics[width=0.9\linewidth]{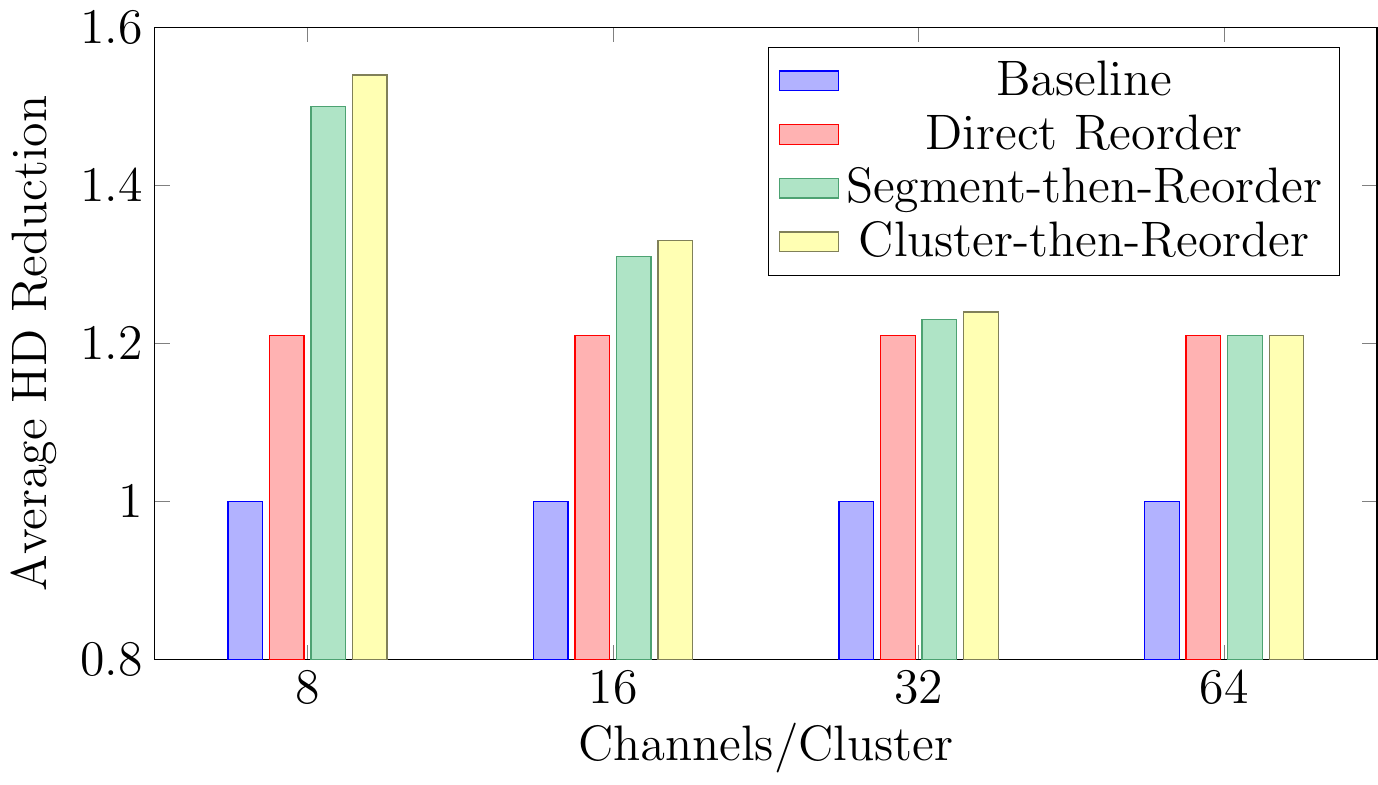}\label{fig:post_train_resnet}}
    \caption{Comparison of the post-training optimization techniques for HD reduction on (a) MobileNetV2 and (b) ResNet26.}
    \label{fig:post_train_compare}
\end{figure}

We first compare the effectiveness of different post-training hamming distance optimization algorithms, including the direct reorder, segment-then-reorder, and cluster-then-reorder algorithms. We select the 1-by-1 convolution layers from the MobileNetV2 and the 3-by-3 layers from the ResNet26 for the evaluation. We compare the hamming distance of different algorithms with the baseline setting without any optimization. As shown in Figure \ref{fig:post_train_compare}, when the number of input channels per cluster is 8, the average hamming distance can be reduced by 1.96$\times$ and 1.54$\times$ for MobileNetV2 and ResNet26, respectively, which translate to 1.62$\times$ and 1.49$\times$ reduction of the average energy consumption.

We also have a more detailed comparison between the segment-then-reorder and the cluster-then-reorder algorithms for MobileNetV2 as shown in Figure~\ref{fig:per_layer_comp}. As shown in the figure, the cluster-then-reorder algorithm usually results in a higher reduction for the even layers, e.g., layer 2, layer 4, etc. These layers are the second 1-by-1 convolution layers in the inverted residual blocks, which have a larger number of input channels and a smaller number of output channels. For these layers, more clusters can be formed to achieve better results. For the even layers with a smaller number of input channels and a larger number of output channels, the two methods perform similarly.


\subsection{Training-Aware Hamming Distance Optimization}

\begin{table*}[!htb]
\def\arraystretch{1.1}
\caption{Average hamming distance and energy reduction of the combined methods (CTR is short for the cluster-then-reorder algorithm).}
\label{tab:combine_comp}
\vskip 0.1in
\centering
\resizebox{0.8\linewidth}{!}{
\begin{sc}
\begin{tabular}{c|cc|cc}
\toprule
                     & \begin{tabular}[c]{@{}l@{}}Best-Layer HD\\ Reduction\end{tabular}  & \begin{tabular}[c]{@{}l@{}}Average HD\\ Reduction\end{tabular} & \begin{tabular}[c]{@{}l@{}}Best-Layer Energy\\ Reduction\end{tabular} & \begin{tabular}[c]{@{}l@{}}Average Energy\\ Reduction\end{tabular} \\
\midrule
Baseline                          &  1.0$\times$     &  1.0$\times$     &  1.0$\times$     & 1.0$\times$    \\
$\lambda=0$, CTR                  & 2.44$\times$     &  1.96$\times$    &  2.27$\times$    &  1.84$\times$       \\
$\lambda=7 \times 10^{-5}$        & 5.95$\times$     &  2.00$\times$    &  4.92$\times$    &  1.86$\times$       \\  
\rowcolor{black!10}                                                      
$\lambda=7 \times 10^{-5}$, CTR   & 10.2$\times$     &  3.79$\times$    &  8.51$\times$    &  2.85$\times$       \\                             
\bottomrule
\end{tabular}
\end{sc}
}
\end{table*}

We now evaluate the effectiveness of the training-aware hamming distance algorithms. We select MobileNetV2 and train the network on Cifar10 and Cifar100 datasets. By controlling the regularization coefficients $\lambda_2$, we explore the trade-off between the accuracy and the reduction of hamming distance. For practical purpose, we constrain the accuracy degradation within 1\%. As shown in Table~\ref{tab:training_aware}, on Cifar10 dataset, the average hamming distance can be reduced by 7.55$\times$, which leads to 6.63$\times$ reduction of the average energy across layers. On Cifar100 dataset, the average hamming distance reduction and the average energy reduction are 2.00$\times$ and 1.86$\times$, respectively.

\subsection{Combined Hamming Distance Optimization.}

We now combine the post-training optimization techniques with the training-aware optimization algorithm. As shown in Table \ref{tab:combine_comp}, the proposed training-aware and post-training optimization techniques can work orthogonal to each other. By combining these optimization techniques, for MobileNetV2 trained on Cifar100, the average hamming distance of streaming the weight matrices can be reduced by 3.79$\times$ and the average datapath energy can be reduced by 2.85$\times$. 

\section{Conclusion}
Energy consumption of arithmetic datapath in a neural network accelerator is heavily dependent on the hamming distance of the input sequence. With the proposed Hamming-Distance-Aware training and post-processing algorithm, the energy consumption of datapath can be significantly reduced. Evaluation with MobileNetV2 and ResNet neural networks shows that our proposed methods can achieve 2.85$\times$ datapath energy reduction on average and up to 8.51$\times$ datapath energy reduction for certain network layers, which demonstrates significant potential in energy-critical neural network accelerator designs.
\label{sec:conclusion}

\bibliography{reference_ml.bib}
\bibliographystyle{sysml2019}

\newpage
\onecolumn

\appendix
\gdef\thesection{Appendix \Alph{section}}


\section{Layer shapes of MobileNetV2 and ResNet26}
\label{sec:layer_shapes_app}

\begin{table*}[!htb]
\caption{Layer shapes of 1-by-1 convolutions in MobileNetV2.}
\label{tab:layer_shape}
\centering
\resizebox{\linewidth}{!}{
\begin{tabular}{c|cccccccccccccccccccccccccccccccccccc}
\toprule
\# Layer & 1 & 2 & 3 & 4 & 5 & 6 & 7 & 8 & 9 & 10 & 11 & 12 & 13 & 14 & 15 & 16 & 17 & 18 & 19 & 20 & 21 & 22 & 23 & 24 & 25 & 26 & 27 & 28 & 29 & 30 & 31 & 32 & 33 \\
\midrule
$C$ & 16 & 96 & 24 & 144 & 24 & 144 & 32 & 192 & 32 & 192 & 32 & 192 & 64 & 384 & 64 & 384 & 64 & 384 & 64 & 384 & 96 & 576 & 96 & 576 & 96 & 576 & 160 & 960 & 160 & 960 & 160 & 960 & 320 \\ 
$K$ & 96 & 24 & 144 & 24 & 144 & 32 & 192 & 32 & 192 & 32 & 192 & 64 & 384 & 64 & 384 & 64 & 384 & 64 & 384 & 96 & 576 & 96 & 576 & 96 & 576 & 160 & 960 & 160 & 960 & 160 & 960 & 320 & 1280 \\
\bottomrule
\end{tabular}
}
\end{table*}

\begin{table*}[!htb]
\caption{Layer shapes of 3-by-3 convolutions in ResNet26.}
\label{tab:layer_shape_app}
\centering
\resizebox{0.9\linewidth}{!}{
\begin{tabular}{c|cccccccccccccccccccccccccccccccccccc}
\toprule
\# Layer & 1 & 2 & 3 & 4 & 5 & 6 & 7 & 8 & 9 & 10 & 11 & 12 & 13 & 14 & 15 & 16 & 17 & 18 & 19 & 20 & 21 & 22 & 23 & 24 & 25 & 26 \\
\midrule
$C$ & 16 & 16 & 16 & 16 & 16 & 16 & 16 & 16 & 16 & 32 & 16 & 32 & 32 & 32 & 32 & 32 & 32 & 32 & 64 & 32 & 32 & 32 & 32 & 32 & 32 & 32 \\
$K$ & 16 & 16 & 16 & 16 & 16 & 16 & 16 & 16 & 32 & 32 & 16 & 32 & 32 & 32 & 32 & 32 & 32 & 64 & 64 & 64 & 64 & 64 & 64 & 64 & 64 & 64 \\
\bottomrule
\end{tabular}
}
\end{table*}

\end{document}